\documentclass{article}

\usepackage{arxiv}

\usepackage[utf8]{inputenc} 
\usepackage[T1]{fontenc}    
\usepackage{hyperref}       
\usepackage{url}            
\usepackage{booktabs}       
\usepackage{amsfonts}       
\usepackage{nicefrac}       
\usepackage{microtype}      
\usepackage{graphicx}
\usepackage{doi}
\usepackage{amsmath,amssymb,latexsym, amscd}
\usepackage{exscale, cite, eps fig, graphics}
\usepackage{array}
\usepackage{float}
\restylefloat{table}
\usepackage{placeins}
\usepackage{longtable}
\usepackage{makecell}
\usepackage[utf8]{inputenc}
\usepackage{graphicx}
\usepackage{caption}
\usepackage{subcaption}
\usepackage{environ}
\usepackage{mathtools}
\usepackage{booktabs}
\usepackage{multirow}
\usepackage{algorithmic}
\usepackage{graphicx}
\usepackage{textcomp}
\usepackage{xcolor}
\usepackage{amsthm}

\title{Intensity Prediction of Tropical Cyclones using Long Short-Term Memory Network}

\author{ \href{https://orcid.org/0000-0002-9818-8966}{\includegraphics[scale=0.06]{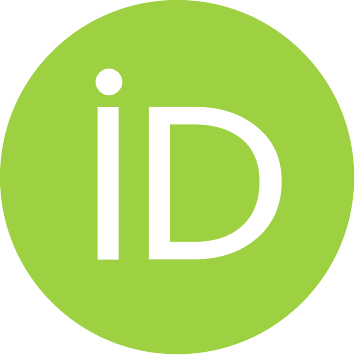}\hspace{1mm}Koushik Biswas} \\
	Department of Computer Science, IIIT Delhi\\
	New Delhi, India, 110020. \\
	\texttt{koushikb@iiitd.ac.in} \\
	\And
	\href{https://orcid.org/0000-0002-5464-929X}{\includegraphics[scale=0.06]{orcid.pdf}\hspace{1mm}Sandeep Kumar} \\
	Department of Computer Science, IIIT Delhi\\
	\&\\
	Department of Mathematics,\\ Shaheed Bhagat Singh College, University of Delhi\\
	New Delhi, India, 110020.\\
	\texttt{sandeepk@iiitd.ac.in, sandeep\_kumar@sbs.du.ac.in} \\
	\AND
	 Ashish Kumar Pandey\\
	Department of Mathematics, IIIT Delhi\\
	New Delhi, India, 110020.\\
	\texttt{ashish.pandey@iiitd.ac.in } \\
}

\begin{document}
\maketitle

\begin{abstract}
Tropical cyclones can be of varied intensity and cause a huge loss of lives and property if the intensity is high enough. Therefore, the prediction of the intensity of tropical cyclones advance in time is of utmost importance. We propose a novel stacked bidirectional long short-term memory network (BiLSTM) based model architecture to predict the intensity of a tropical cyclone in terms of Maximum surface sustained wind speed (MSWS). The proposed model can predict MSWS well advance in time (up to 72 h) with very high accuracy. We have applied the model on tropical cyclones in the North Indian Ocean from 1982 to 2018 and checked its performance on two recent tropical cyclones, namely, Fani and Vayu. The model predicts MSWS (in knots) for the next 3, 12, 24, 36, 48, 60, and 72 hours with a mean absolute error of 1.52, 3.66, 5.88, 7.42, 8.96, 10.15, and 11.92, respectively.  
\end{abstract}


%

\section{Introduction}

The weather-related forecast is one of the difficult problems to solve due to the complex interplay between various cause factors. Accurate tropical cyclone intensity prediction is one such problem that has huge importance due to its vast social and economic impact. Cyclones are one of the devastating natural phenomena that frequently occur in tropical regions. Being a tropical region, Indian coastal regions are frequently affected by tropical cyclones \cite{chaudhuri2012appraisal} that originate into the Arabian Sea (AS)  and Bay of Bengal (BOB), which are parts of the North Indian Ocean (NIO). With the increasing frequency of cyclones in NIO \cite{frank1999effects}, it becomes more crucial to develop a model that can forecast the intensity of a cyclone for a longer period of time by observing the cyclone only for a small period of time. Various statistical and numerical methods have been developed to predict the intensity of cyclones \cite{jarvinen1979statistical, demaria1999updated, baik1998tropical, chaudhuri2009severity, dvorak1984tropical} but all these methods lack effectiveness in terms of accuracy and computation time. 

India Meteorological Department (IMD) keep track of tropical cyclones which originates in the North Indian Ocean between $100^{\circ}$E and $45^{\circ}$E. Typically, the intensity of the tropical cyclone is stated in terms of "Grade", which is directly derived from different ranges of the Maximum Sustained Surface Wind Speed (MSWS), see Table~\ref{tab:grade}~\cite{imd}. Therefore, MSWS being a continuous variable is a better choice for the intensity prediction \cite{chaudhuri2017swarm}, which we adopt for this work. We have used latitude, longitude, MSWS, estimated central pressure, distance, direction, and sea surface temperature as input features.
\begin{table}[!h]
    \centering
    \caption{The classification of the low pressure systems by IMD.}
    \label{tab:grade}
    \begin{tabular}{c|c|c} \hline
     {\bf Grade} & {\bf Low pressure system}   &  {\bf MSWS (in knots) }\\ \hline
         0 & Low Pressure Area (LP) & $<$17\\
1 & Depression (D) & 17-27\\
2 & Deep Depression (DD) & 28-33\\ 
3 & Cyclonic Storm (CS) & 34-47\\ 
4 & Severe Cyclonic Storm (SCS) & 48-63\\ 
5 & Very Severe Cyclonic Storm (VSCS) & 64-89\\ 
6 & Extremely Severe Cyclonic
Storm (ESCS) & 90-119\\
7 & Super Cyclonic Storm (SS) & $\geq$120\\  \hline
    \end{tabular}
\end{table}

Recently, Artificial neural Networks (ANNs) have been successful in capturing the complex non-linear relationship between input and output variables \cite{APHY2012, ASSA2012, KRC2009}. ANNs are also explored recently to predict the cyclone intensity \cite{chaudhuri2015track, chaudhuri2017swarm, roy2012tropical, mohapatra2013evaluation}. These studies use the various recordings of weather conditions for a particular time point and predict the intensity of cyclones at a particular future time point. However, this does not allow us to fully utilise the time series data available. Instead, we use the Long Term Short Memory (LSTM) network to forecast the intensity of tropical cyclones in NIO. By using LSTM, we are able to use the different weather characteristics for a certain number of continuous-time points to forecast the cyclone intensity for an immediately succeeding a certain number of time points. 

In related works \cite{chaudhuri2015track, chaudhuri2017swarm, mohapatra2013evaluation} for tropical cyclones in NIO, the MSWS has been predicted for a particular future time point. Although it is important to know the intensity at a particular time point, the change in intensity as the tropical cyclone progresses is equally important to know. As mentioned earlier, using ANNs, this continuous variation can not be captured effectively. Our work is unique in the sense that we have used an LSTM based model for the first time to predict the cyclone intensity for multiple successive future time points and report the combined accuracy for these time points. The reported combined accuracy outperforms the accuracy reported at a single point in other works. Our model works consistently well, even for a large number of future time points, and increases gradually with the number of future time points. In Section II, we present a brief description of our model, and in Section III, the used dataset is described. Section IV presents results and their analysis, and Section V includes the conclusion and future directions.  


\section{METHODOLOGY}
\subsection{Artificial Neural Networks}
\begin{figure}[H]
    \centering
    \includegraphics[width =2in]{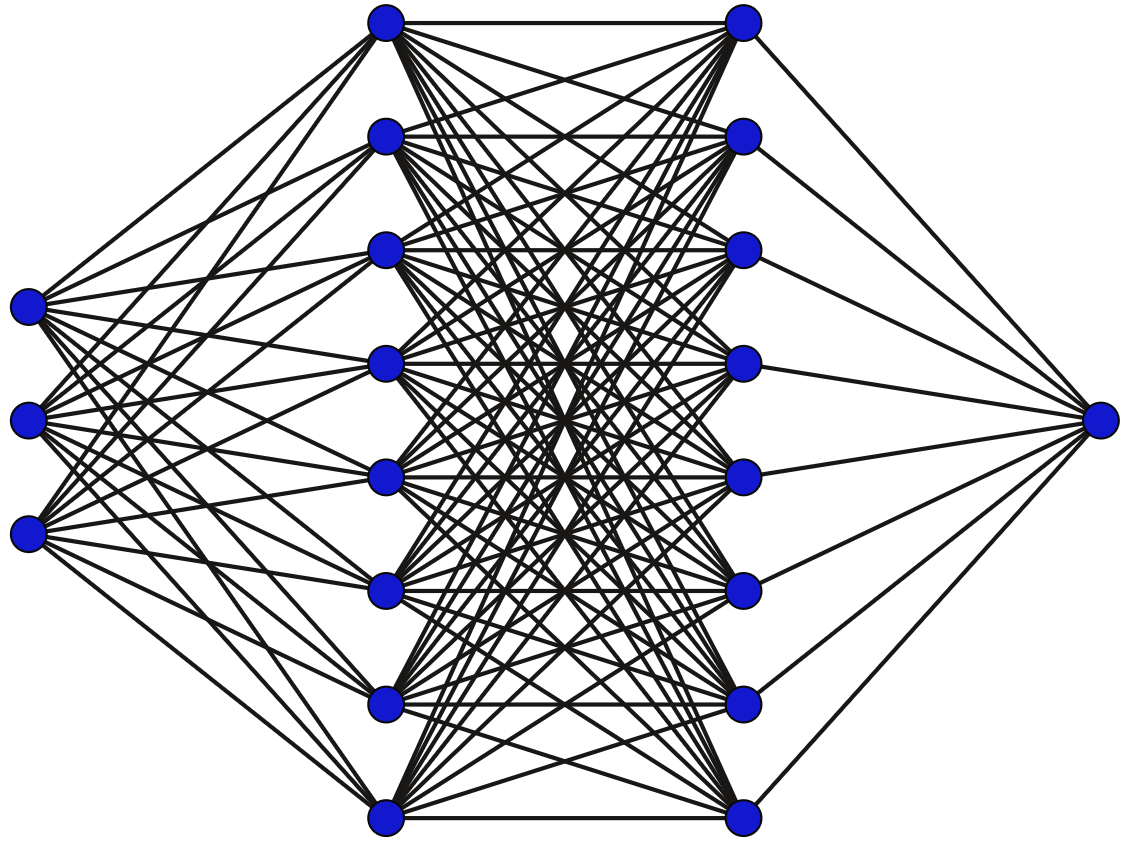}
    \caption{Fully Connected ANN}
    \label{figla1}
\end{figure}

An artificial neural network (ANN) is a connected network inspired by the human brain's biological neurons. ANN has an input layer, an output layer, and multiple hidden layers. Each hidden layer contains several artificial neurons at which incoming information is first linearly combined using weights and bias and then acted upon by an activation function. Mathematically, for input matrix $X$, the $i$th neuron in the $j$th layer can be written as 
\begin{align}
    h_j^i = \sigma(W_j^iX + b_j^i) 
\end{align}
where $W_j^i$ and $b_j^i$ are weight matrix and bias vector of the corresponding neuron, respectively, and $ \sigma $ is the activation function. The weights and bias at each neuron are updated using the gradient descent algorithm to make the final loss as small as possible. An example of a fully connected ANN with two hidden layers is given in the Figure~\ref{figla1}. A type of ANN that can handle time series or sequential data is Recurrent Neural Network, which we will discuss next.

\subsection{Recurrent Neural Network} 
\begin{figure}[H]
    \centering
    \includegraphics[width = 3in]{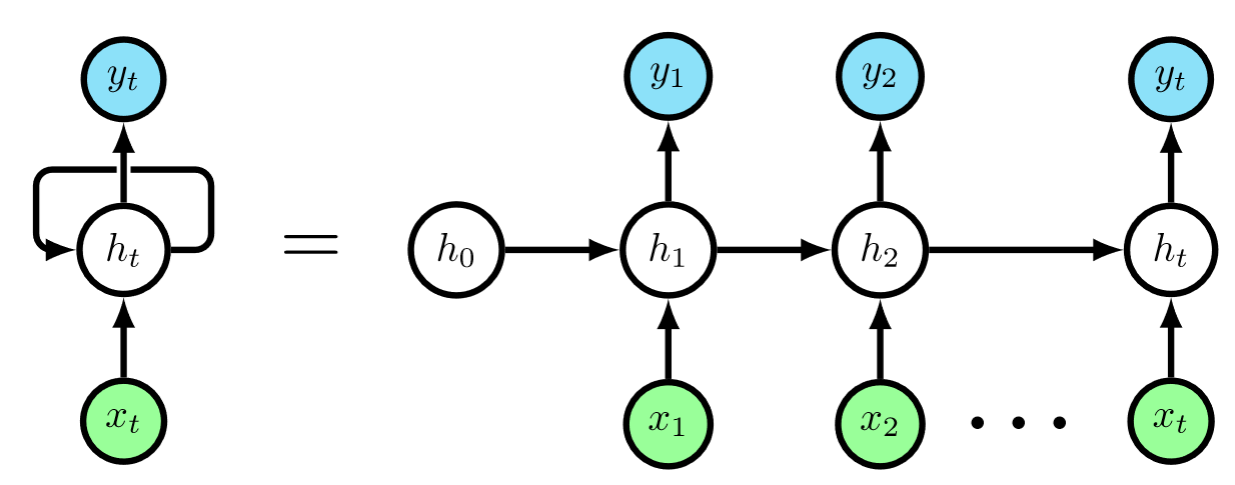}
    \caption{Recurrent Neural Network}
    \label{fig2}
\end{figure}

Recurrent Neural Networks (RNNs) can take a sequence of inputs and produce a sequence of outputs and the outputs are influenced not just by weights applied on inputs like a regular ANN, but also by a hidden state vector representing the learned information based on prior inputs and outputs \cite{rnn1, rnn2, rnn3, rnn4}. RNN can be represented by a chain-like structure, see Figure~\ref{fig2}, where the lower, middle, and upper chains represent the sequence of inputs, hidden state vector, and sequence of outputs, respectively. Mathematically, a simple RNN can be written as 
\begin{align}
    h_t=&\sigma(W_hx_t+U_hh_{t-1}+b_h)  \\
 y_t=& \sigma(W_yh_t+b_y) 
\end{align}
where $\sigma$ is the activation function, $x_t$ is the input vector at timestamp $t$, $h_t$ is the hidden state vector at timestamp $t$, $y_t$ is the output vector at timestamp $t$, $h_{t-1}$ is the output vector at timestamp ${(t-1)}$, $W_h, W_y$ and $U_h$ are weight matrices, and $b_h, b_y$ are the biases. 
The gradient vector of RNN can increase or decrease exponentially during the training period, which leads to exploding or vanishing gradient problems because of which an RNN cannot retain a very long term history from the past. This problem is solved by Long Short Term Memory Networks. 

\subsection{Long Short Term Memory Networks}
\begin{figure}[H]
    \centering
    \includegraphics[width=4in,height=4in,keepaspectratio]{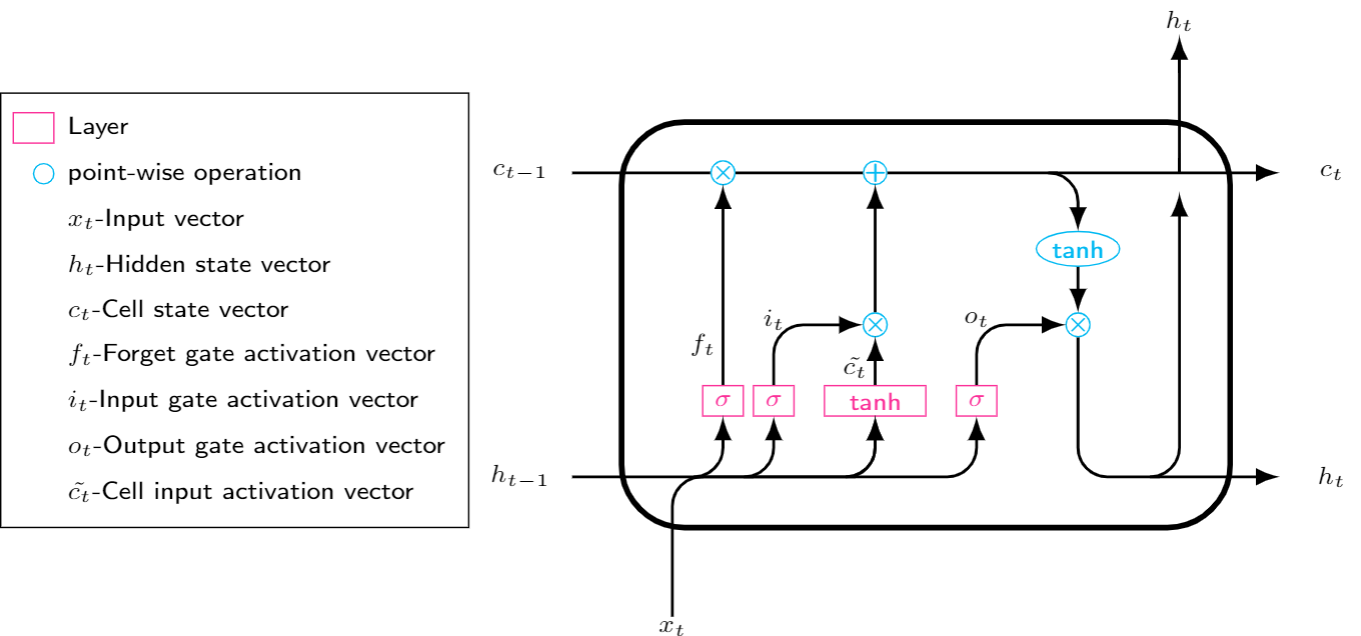}
    \caption{Example of an LSTM Cell}
    \label{figla23}
\end{figure}

Long Short Term Memory Network (LSTM) was first introduced in 1997 by Hochreiter and Schmidhuber\cite{10.1162/neco.1997.9.8.1735} and further few improved version of LSTM was proposed later \cite{lstm1, lstm2, lstm3}. LSTM was defined in such a way so that it can remember long term dependencies from past and overcome the issue of vanishing and gradient problem of RNN.  An LSTM network works on the philosophy of  selectively forget, selectively read, and  selectively write. LSTM can add or delete information to the cell state. This states are called gates. LSTM has three gates which are usually known as Input, Forget and Output gate. The equations for the three LSTM gates are
\begin{align}
f_t = &\sigma (w_f.x_t+u_f.h_{t-1}+b_f) \\ i_t = &\sigma (w_i.x_t+u_i.h_{t-1}+b_i) \\
o_t = &\sigma (w_o.x_t+u_o.h_{t-1}+b_o)
\end{align}
and the equations for the cell state, candidate cell state and output are
\begin{align}
   \tilde{c_{t}} = &\sigma (w_c.x_t+u_c.h_{t-1}+b_c)\\ 
   c_t =& f_t.c_{t-1} +  i_t.c_t \tag{8}\\
  h_t =& o_t.\sigma(c_t) 
\end{align}
where $i_t$ represents input gate, $f_t$ represents forget gate, $o_t$ represents output gate, $h_{t-1}$ is output from previous LSTM block at $(t-1)$ timestamp, $x_t$ is input at current timestamp, $c_t$ represents cell state at timestamp $t$, $\tilde{c_t}$ is candidate for cell state at timestamp $t$, and $w_f, w_i, w_c, w_o, u_f, u_i, u_c, u_o$ are weight matrices, $b_f, b_i, b_c, b_o$ are bias vectors and $\sigma$ is the activation function, see figure~\ref{figla23}.

\subsection{Stacked LSTM}
Stacked LSTM is an extended version of the LSTM model. In this model, there are multiple hidden layers where the next layer is stacked on top of the previous layer, and each layer contains multiple LSTM cells. Stacking layers make the model more deeper and help to learn patterns in sequence-learning and time-series problems more accurately.

\subsection{Bidirectional LSTM}A more modified and classic version of LSTM is Bidirectional LSTM (BiLSTM) \cite{BiLstm}. LSTM is trained to learn in one direction; on the other hand, BiLSTM learns in two directions, one from past to future and another from future to past. BiLSTM has two separate LSTM layers in opposite directions of each other. The input sequence is fed into one layer in the forward direction and another layer in the backward direction. Both of the layers are connected to the same output layer, and it collects information from the past and future simultaneously. 

\subsection{Dropout}
A model will learn complex and dynamical behaviour more accurately if hidden units are independent of each other while learning features. It has been observed that in neural network models, some of the neurons become highly correlated and dependent on each other while the rest are independent. They can significantly affect model performance, which may lead to overfitting. This is called co-adaptation, and it's a major issue in large networks. This problem is solved using the dropout method \cite{dropout}. Dropout is a regularisation technique used in neural network models to prevent the model from overfitting.  It randomly ignores neurons with probability $1-p$ and keeps neurons with probability $p$ during training time. It helps the model to learn more powerful features and patterns from data.

\section{Data}
Various organisations around the world keep a record of all tropical cyclones in that region; such records are generally known as Best Track Datasets (BTDs).  In this study, we have used the BTD of tropical cyclones in the North Indian ocean provided by the Regional Specialized Meteorological Centre, New Delhi \footnote{\url{http://www.rsmcnewdelhi.imd.gov.in/index.php?option=com_content&view=article&id=48&Itemid=194&lang=en}}. 

\begin{figure}[!htbp]
\centering
\subfloat[Cyclones from 1982-2018]{\includegraphics[width=5.5in,height=5in,keepaspectratio]{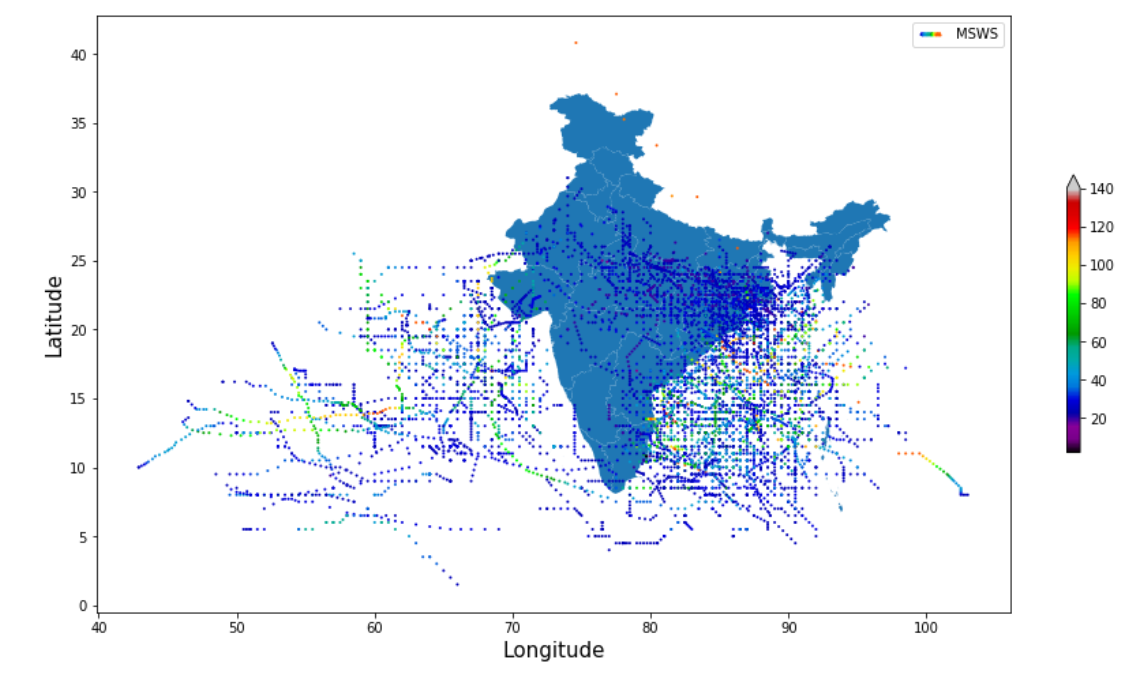}
\label{figla2a}}
\hfill
\subfloat[Vayu Cyclone, 2019]{\includegraphics[width=3.3in,height=3.5in,keepaspectratio]{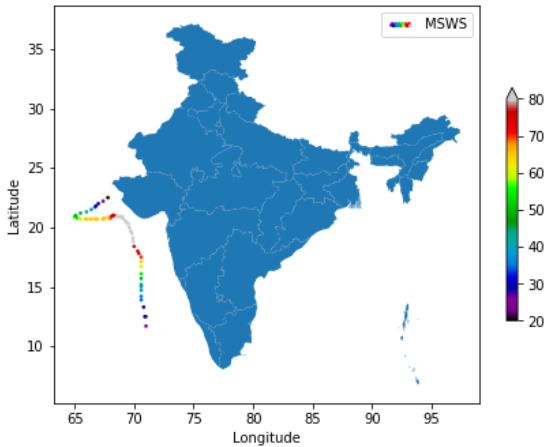}
\label{figla2b}}
\hfill
\subfloat[Fani  Cyclone, 2019]{\includegraphics[width=2.8in,height=2.7in,keepaspectratio]{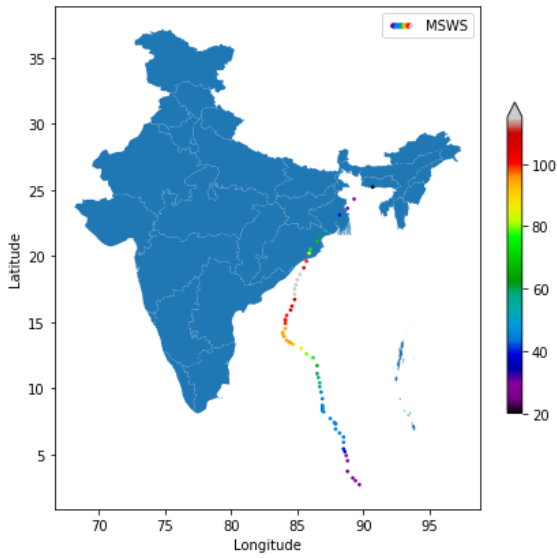}
\label{figla2c}}
\caption{North  Indian  ocean  cyclone’s trajectory with MSWS}
\end{figure}

The dataset contains three hourly records of 341 tropical cyclones from 1982 to 2018 in the NIO. 
There are a total of 7662 recordings, and each recording contains information about the cyclone's basin of origin (AS or BOB), name (if any), date and time of occurrence, latitude, longitude, estimated central pressure (ECP), and MSWS, pressure drop (PD), T.No., and grade. The trajectories (with MSWS) of all the cyclones in NIO along with the trajectory (with MSWS) of two recent devastating cyclones Vayu and Fani are shown in the Figures \ref{figla2a}, \ref{figla2b}, \ref{figla2c} respectively.  

After processing the dataset for possible errors, we obtained a dataset of 341 cyclones with an average number of 27 recordings. The largest cyclone has 90 such recordings. The dataset has lots of missing data, and it is handled using imputing techniques. We have used pandas linear interpolation techniques to fill the missing values.

We have used latitude, longitude, ECP, and MSWS from the BTD. The sea surface temperature (SST) is an important factor for cyclone generation and governs its intensity. SST is obtained from the NOAA dataset provided at \footnote{\url{http://apdrc.soest.hawaii.edu/erddap/griddap/hawaii_soest_afc8_9785_907e.html}}. We have generated two new features, distance and direction from latitudes and longitudes \footnote{\url{https://www.movable-type.co.uk/scripts/latlong.html}}. The features latitude, longitude, MSWS, ECP, distance, direction, and SST are used in our model as input variables.


It has been observed that RNN learns from its inputs equally with the scaled data. We have kept MSWS in original scale and  re-scaled latitude, longitude, ECP, SST, distance and direction to [-1,1] range. We have utilized Scikit learn for features re-scaling using MinMaxScaler transformation \cite{scikit-learn}. MinMaxScaler transformation will map the interval $[\min, \max]$ one-to-one to the interval $[a,b]$, defined as follows
\begin{align}
    f(x) = a + \left(\frac{b-a}{\max-\min}\right)(x-\min) 
\end{align}
with the assumption that $\min \neq \max$ and $f(\min) = a$ and $f(\max) = b$, and $f$ is the transformation.

\section{Training and Proposed Model implementation}

\subsection{Model training}
Model weights and biases have been trained and updated via back-propagation algorithm. We consider Mean Square Error (MSE) as the loss function and evaluate the model performance using Root Mean Square Error (RMSE) and Mean Absolute Error (MAE). The definitions are given as follow
\begin{align}
    \operatorname{MSE} = \frac{1}{n}\sum_{i=1}^{n}(y_{i} - \bar{y_{i}})^{2} \\
\operatorname{RMSE} = \sqrt{\frac{1}{n}\sum_{i=1}^{n}(y_{i} - \bar{y_{i}})^{2}} \\
\operatorname{MAE} = \frac{1}{n}\sum_{i=1}^{n}\left | y_{i} - \bar{y_{i}} \right |
\end{align}
where $y_i$ is the actual value and $\bar{y_i}$ is the model predicted value.

\subsection{Model Implementation}
\begin{figure}[H]
    \centering
    \includegraphics[width=4.7in,height=4.7in,keepaspectratio]{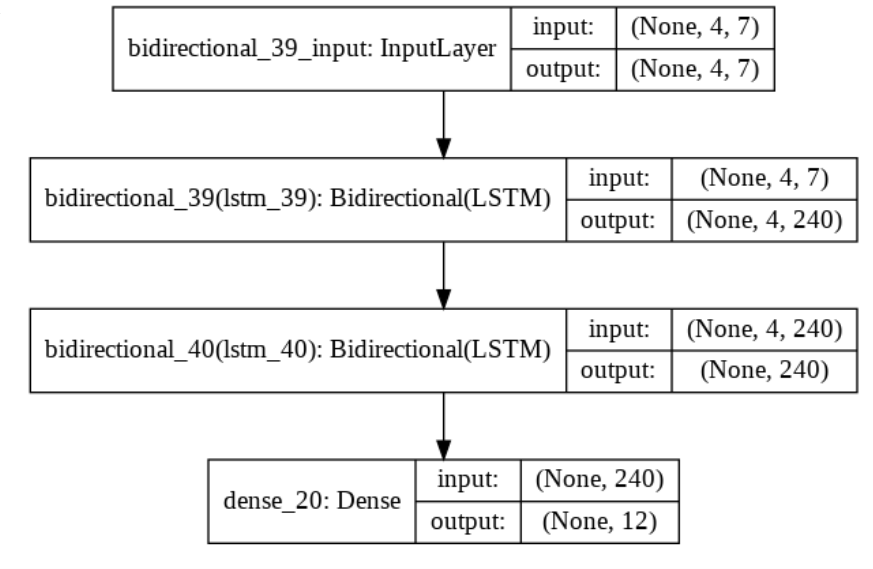}
    \caption{Our Proposed BiLSTM model for $T_1=4$ and $T_2=12$}.
    \label{figla2}
\end{figure}

We have considered a total of four stacked BiLSTM layers (one input, one output, and two hidden layers) for our proposed model architecture. Latitude, longitude, distance, direction, SST, ECP, and MSWS used as data tuples in the input layer. 
We have implemented our proposed model using Keras API \cite{chollet2015keras}. Keras is an API which runs on top of low-level language TensorFlow \cite{tensorflow2015-whitepaper}, developed by Google. We use a learning rate 0.01, which is useful to update weights at each layer of BiLSTM and is improved using Adaptive moment estimation (Adam) \cite{adam} and it helps to minimise the loss function. We have set dropout value equals to 0.02 in internal layers. Our proposed model is shown in Figure~\ref{figla2}. It has been generated using Keras API. 


\begin{table*}[!t]
\caption{Our Proposed BiLSTM Model performance on Indian ocean cyclone data.
\label{tabel1}}
\begin{center}
  \begin{tabular}{| p{1.8cm} |p{1.8cm}|p{1.8cm} | p{1.4cm} | p{1.2cm} | p{1.2cm}| p{1.4cm} | p{1.2cm} | p{1.2cm}|}
    \hline
    \multicolumn{3}{|c|}{Data Used} & \multicolumn{3}{c|}{Mean Absolute Error} & \multicolumn{3}{c|}{Root Mean Square Error} \\ 
    \hline
    Training Points $T_1$ (hours)  & Predict Points $T_2$ (hours)  & Training data size (no of cyclones) & 5-fold MAE on Validation Data &   Test on Cyclone Vayu & Test on Cyclone Fani & 5-fold RMSE on Validation Data &  Test on Cyclone Vayu & Test on Cyclone Fani  \\ \hline
    \multirow{7}{*}{4 (12)} & 1 (3)&  7211 (320) & 1.54 & 3.95 & 3.20 & 3.12  & 4.47 & 4.51\\ 
    \cline{2-9}
     & 4 (12) & 6263 (306) & 3.66 & 9.81  &  7.12 & 6.72 & 11.52 &  9.17 \\
    \cline{2-9}
     & 8 (24) & 5109 (266) & 5.88 & 11.31  & 13.52  & 9.95 & 13.98 & 18.48  \\
    \cline{2-9}
    & 12 (36) & 4110 (228) & 7.42 & 17.52  &  20.04 & 12.64 & 21.03 & 26.05  \\
    \cline{2-9}
    & 16 (48) & 3244 (195) & 8.96 &  15.50 & 21.32 & 14.82 & 18.22 & 27.12   \\
    \cline{2-9}
    & 20 (60) & 2512 (165) & 10.15 & 16.98 & 22.72 & 16.07 & 21.34  &  28.87  \\
    \cline{2-9}
    & 24 (72) & 1888 (138) & 11.92 &  15.01 & 22.97 & 17.87 & 20.83 & 28.14 \\
    \cline{1-9}
    \multirow{7}{*}{6 (18)} & 1 (3)& 6574 (311) & 1.55 & 2.98 & 2.32 & 3.20 & 3.31 &  3.56 \\ 
    \cline{2-9}
    & 4 (12) & 5657 (277) & 3.72 & 8.01  & 7.03 & 6.37 & 10.36 & 9.42   \\
    \cline{2-9}
     & 8 (24) & 4588 (243) & 6.19 & 13.24  & 12.53 & 10.62 & 15.84 & 16.23   \\
    \cline{2-9}
    & 12 (36) & 3657 (211) & 7.92  &  17.32 & 20.54 & 13.23 & 20.01 &  26.12  \\
    \cline{2-9}
     & 16 (48) & 2864 (179) & 9.31 & 10.82 & 27.74 & 15.35  & 13.05  & 27.52     \\
     \cline{2-9}
     & 20 (60) & 2185 (151) & 10.22 & 11.82 & 22.56 &  16.42 & 15.01  &  28.52    \\
    \cline{2-9}
     & 24 (72) &  1627 (117) & 11.52 & 16.34 & 19.82 & 17.85 & 28.98  & 35.86    \\
     \cline{1-9}
    \multirow{7}{*}{8 (24)} & 1 (3)&  5957 (300) & 1.52 & 2.32 & 2.52 & 3.32 & 2.98 & 3.56 \\ 
    \cline{2-9}
    & 4 (12) & 5109 (266) & 3.82 & 7.08  & 6.82 & 6.88 & 9.01 & 9.02   \\
    \cline{2-9}
    & 8 (24) & 4110 (228) & 5.98 &  9.92 & 13.46 & 10.82 & 12.32 &  18.57  \\
    \cline{2-9}
    & 12 (36) & 3244 (195) & 8.12 & 10.66  & 21.62 & 13.18 & 13.03  & 30.05   \\
    \cline{2-9}
     & 16 (48) & 2512 (165) & 9.91 & 10.02  & 25.11 & 15.82 & 11.82 & 30.98   \\
     \cline{2-9}
     & 20 (60) & 1888 (138) & 11.52 & 14.02  & 27.12 & 17.56 & 16.52 & 33.26    \\
    \cline{2-9}
     & 24 (72) & 1399 (104) & 12.01 & 18.82   & 32.52 & 18.52 & 21.98 & 39.80   \\
     
    \cline{1-9}
    \multirow{7}{5em}{12 (36)} & 1 (3)& 4843 (255) & 1.72  &  2.22 & 3.45 & 3.44 & 2.52 & 5.41   \\ 
    \cline{2-9}
    & 4 (12) & 4110 (228) & 3.96 &  4.52  & 5.37 & 7.52 & 6.01  &  8.12 \\ 
    \cline{2-9}
     & 8 (24) & 3244 (195) & 6.52 & 8.37   & 17.85 & 11.32 & 9.75 & 20.33  \\ \cline{2-9}
     \cline{2-9}
     & 12 (36) & 2512 (165) & 8.62 & 8.01   & 21.10 & 14.31 & 10.51 & 27.33  \\ 
     \cline{2-9}
     & 16 (48) & 1888 (138) & 10.32 & 8.51   & 29.31 & 16.57 & 10.85 & 27.82     \\
    \cline{2-9}
    & 20 (60) & 1399 (104) & 10.98 & 14.02   &  21.27 & 17.63 & 17.07 &  28.02    \\
    \cline{2-9}
     & 24 (72) &  1016 (80) & 12.01 & 15.55   & 34.34 & 18.57  & 19.52 &  39.37 \\
    \cline{1-9}
  \end{tabular}
\end{center}
\end{table*}
\begin{figure}[!htbp]
\centering
\subfloat[$T_1$ = 4 and $T_2$ = 12]{\includegraphics[width=6.2in,height=6in,keepaspectratio]{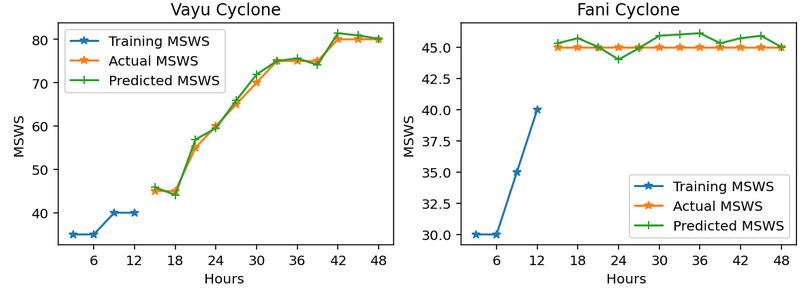}
\label{fig_first_case}}
\hfil
\subfloat[$T_1$ = 6 and $T_2$ = 16]{\includegraphics[width=6.2in,height=6in,keepaspectratio]{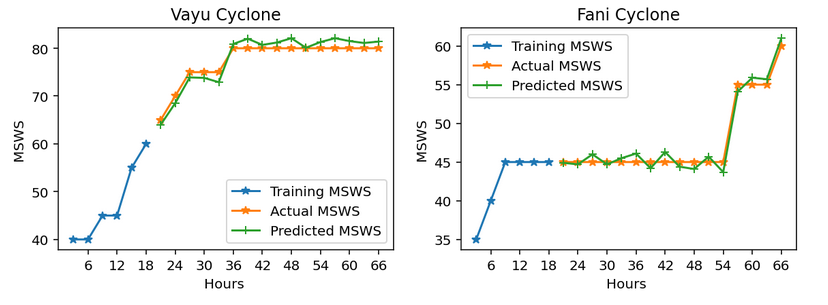}
\label{fig_second_case}}
\hfil
\subfloat[$T_1$ = 8 and $T_2$ = 20]{\includegraphics[width=6.2in,height=6.9in,keepaspectratio]{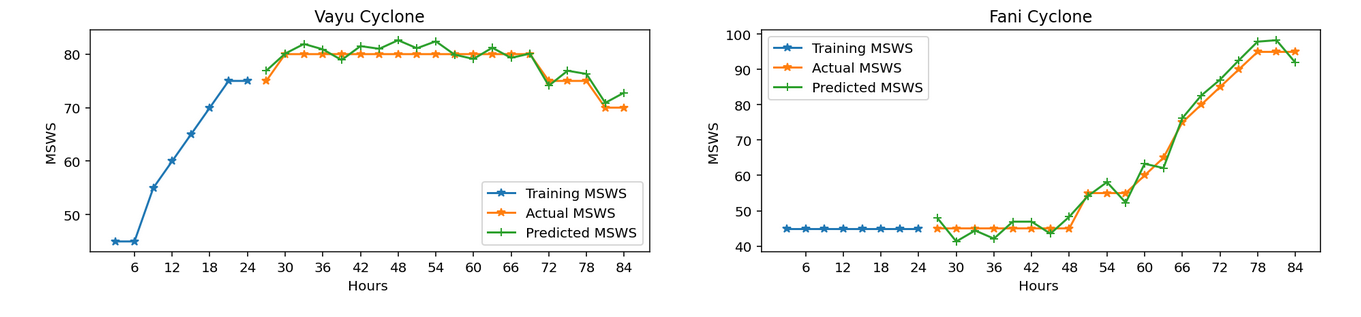}
\label{fig_third_case}}
\hfil
\subfloat[$T_1$ = 12 and $T_2$ = 24]{\includegraphics[width=6in,height=7.2in,keepaspectratio]{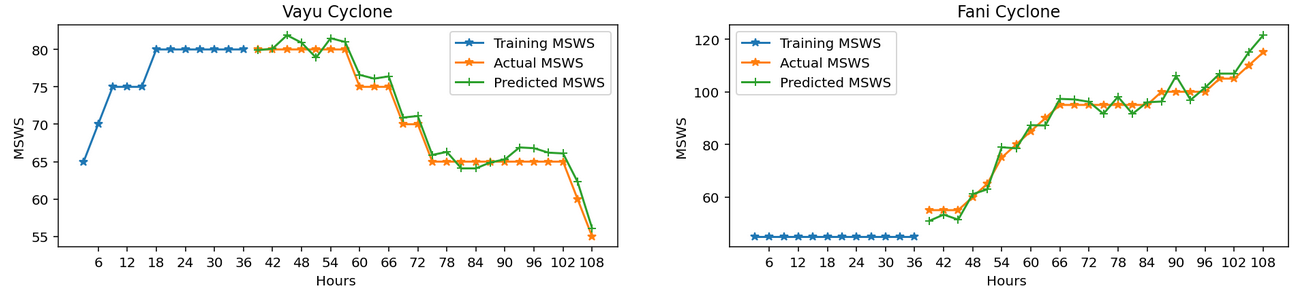}
\label{fig_fourth_case}}
\caption{Actual and Model predicted MSWS for Testing Data- Cyclone Vayu and Fani.}
\label{fig_sim}
\end{figure}


\section{Result and Analysis}

In this study, we are interested in predicting a cyclone's intensity in the form of MSWS for $T_2$ number of time points at a regular interval, using the cyclone information for $T_1$ number of time points, again given at regular time interval. For example, if our $T_1$ is 4 and $T_2$ is 6, then we want to train a model that uses a cyclone data of 4 regular time points and predict the intensity of the cyclone immediately succeeding  6 regular time points. We report the results for different combinations of $T_1$ and $T_2$. To compare our results with the existing related works, the model performance is reported in terms of RMSE  and MAE. We have reported our model's performance for two recent cyclones Vayu and Fani, along with 5-fold cross-validation scores in Table~\ref{tabel1}.  These two named cyclones are not part of the training data.

The predicted intensity of cyclones Vayu and Fani for different combinations of $T_1$ and $T_2$ is shown in Figures~\ref{fig_first_case}, ~\ref{fig_second_case}, ~\ref{fig_third_case}, and ~\ref{fig_fourth_case} along with the actual intensity for $T_1 + T_2$ number of time points. These figures show that our model effectively captures the future evolution of cyclone intensity from the given training data. The graphs for Fani cyclones in Figures~\ref{fig_third_case}, and \ref{fig_fourth_case}, show that though the intensity of cyclone remains almost constant during training time, still our model predicts the intensity quite accurately for future $T_2$ number of time points. This demonstrates that our model is effectively learning from its input features latitude, longitude, direction, distance, ECP, and SST. A similar but reverse behaviour can be observed from the graph of the Fani cyclone in Figure \ref{fig_first_case}, where though the intensity is continuously increasing during training time, still, our model accurately captures the almost constant intensity for future $T_2$ number of time points. Thus, we can conclude that our model is learning the complex non-linear relationship of input features with the output effectively and successfully predicts sudden changes in the intensity of cyclones. Graphs, as reported in Figure~\ref{fig_sim} for any future cyclone, can be used for preventives measures well in time.

From Table~\ref{tabel1}, it is clear that our model is able to learn MSWS with good accuracy in terms of MAE and RMSE for different combinations of $T_1$ and $T_2$. If we fix $T_1$, from the Table~\ref{tabel1}, we can see that as $T_2$ increases, error increases as well. This is expected because we are predicting MSWS for a longer duration as $T_2$ increases from the same set of training points. Moreover, as $T_1$ increases, there is no significant difference between errors. For example, if $T_1=4$ and $T_2=1$, MAE is 1.54 and if $T_1=8$ and $T_2=1$, MAE is 1.52. Similar trend can be observed for other combinations of $T_1$ and $T_2$. So, even though we increased the number of training points two-fold, the change in MAE is not significant. This indicates that the model is learning most from the initial few data points. For practical purposes, this is very important as it reduces both the waiting and computational time.

The model can predict MSWS up to 24 h well within 6 knots. From Table~\ref{tab:grade}, range of MSWS for higher Grades ($\geq 3$) is at least 15. It means with a high probability, the model will predict the grades greater than equal to 3 accurately up to 24 h.

\section{Conclusion}

The authors presented a BiLSTM model to predict the intensity of tropical cyclones in NIO for continuous long-duration future time points with high accuracy. The BiLSTM model is used for the first time for the intensity forecast. As we are able to achieve high accuracy for longer period intensity forecast, this means that our model is successfully able to capture the complex relationship between various cause factors behind cyclone formation that governs the evolution of cyclone intensity.  

The most crucial time point to predict the cyclone intensity would be at landfall. In future, one can try to predict the time of landfall and then train a model that accurately predicts the intensity at that instance. This will make the forecast more practical and effective as the government agencies can take more precise preventive steps both in terms of the region to focus and level of preparedness.


\section*{Acknowledgment}
The authors would like to thank the India Meteorological Department (IMD), New Delhi, for
providing BTD for this study. The authors acknowledge NOAA for providing the SST data.
\bibliographystyle{unsrt} 
\bibliography{reference.bib}  

\begin{thebibliography}{10}

\bibitem{chaudhuri2012appraisal}
Sutapa Chaudhuri, Anirban Middey, Sayantika Goswami, and Soumita Banerjee.
\newblock Appraisal of the prevalence of severe tropical storms over indian
  ocean by screening the features of tropical depressions.
\newblock {\em Natural hazards}, 61(2):745--756, 2012.

\bibitem{frank1999effects}
William~M Frank and Elizabeth~A Ritchie.
\newblock Effects of environmental flow upon tropical cyclone structure.
\newblock {\em Monthly weather review}, 127(9):2044--2061, 1999.

\bibitem{jarvinen1979statistical}
Brian~R Jarvinen and Charles~J Neumann.
\newblock Statistical forecasts of tropical cyclone intensity for the north
  atlantic basin.
\newblock 1979.

\bibitem{demaria1999updated}
Mark DeMaria and John Kaplan.
\newblock An updated statistical hurricane intensity prediction scheme (ships)
  for the atlantic and eastern north pacific basins.
\newblock {\em Weather and Forecasting}, 14(3):326--337, 1999.

\bibitem{baik1998tropical}
Jong-Jin Baik and Hong-Sub Hwang.
\newblock Tropical cyclone intensity prediction using regression method and
  neural network.
\newblock {\em Journal of the Meteorological Society of Japan. ser. ii},
  76(5):711--717, 1998.

\bibitem{chaudhuri2009severity}
Sutapa Chaudhuri and Anindita De~Sarkar.
\newblock Severity of tropical cyclones atypical during el nino--a statistical
  elucidation.
\newblock {\em Asian Journal of Water, Environment and Pollution}, 6(4):79--85,
  2009.

\bibitem{dvorak1984tropical}
Vernon~F Dvorak.
\newblock {\em Tropical cyclone intensity analysis using satellite data},
  volume~11.
\newblock US Department of Commerce, National Oceanic and Atmospheric
  Administration~…, 1984.

\bibitem{imd}
India~Meteorological Department(IMD).
\newblock Frequently asked questions on tropical cyclones.
\newblock Available at
  \url{http://www.rsmcnewdelhi.imd.gov.in/images/pdf/cyclone-awareness/terminology/faq.pdf}.

\bibitem{chaudhuri2017swarm}
S~Chaudhuri, D~Basu, D~Das, S~Goswami, and S~Varshney.
\newblock Swarm intelligence and neural nets in forecasting the maximum
  sustained wind speed along the track of tropical cyclones over bay of bengal.
\newblock {\em Natural Hazards}, 87(3):1413--1433, 2017.

\bibitem{APHY2012}
MM~Ali, P.~Jagadeesh, I-I Lin, and Je-Yuan Hsu.
\newblock A neural network approach to estimate tropical cyclone heat potential
  in the indian ocean.
\newblock {\em IEEE Geoscience and Remote Sensing Letters}, 9:1114--1117, 11
  2012.

\bibitem{ASSA2012}
Kumar Abhishek, M.P. Singh, Saswata Ghosh, and Abhishek Anand.
\newblock Weather forecasting model using artificial neural network.
\newblock {\em Procedia Technology}, 4:311 -- 318, 2012.
\newblock 2nd International Conference on Computer, Communication, Control and
  Information Technology( C3IT-2012) on February 25 - 26, 2012.

\bibitem{KRC2009}
Rita Kovordányi and Chandan Roy.
\newblock Cyclone track forecasting based on satellite images using artificial
  neural networks.
\newblock {\em ISPRS Journal of Photogrammetry and Remote Sensing},
  64:513--521, 11 2009.

\bibitem{chaudhuri2015track}
Sutapa Chaudhuri, Debashree Dutta, Sayantika Goswami, and Anirban Middey.
\newblock Track and intensity forecast of tropical cyclones over the north
  indian ocean with multilayer feed forward neural nets.
\newblock {\em Meteorological Applications}, 22(3):563--575, 2015.

\bibitem{roy2012tropical}
Chandan Roy and Rita Kovord{\'a}nyi.
\newblock Tropical cyclone track forecasting techniques―a review.
\newblock {\em Atmospheric research}, 104:40--69, 2012.

\bibitem{mohapatra2013evaluation}
M~Mohapatra, BK~Bandyopadhyay, and DP~Nayak.
\newblock Evaluation of operational tropical cyclone intensity forecasts over
  north indian ocean issued by india meteorological department.
\newblock {\em Natural Hazards}, 68(2):433--451, 2013.

\bibitem{rnn1}
Michael~I. Jordan.
\newblock {\em Attractor Dynamics and Parallelism in a Connectionist Sequential
  Machine}, page 112–127.
\newblock IEEE Press, 1990.

\bibitem{rnn2}
D.~E. {Rumelhart} and J.~L. {McClelland}.
\newblock {\em Learning Internal Representations by Error Propagation}, pages
  318--362.
\newblock 1987.

\bibitem{rnn3}
Axel Cleeremans, David Servan-Schreiber, and James~L. McClelland.
\newblock Finite state automata and simple recurrent networks.
\newblock {\em Neural Comput.}, 1(3):372–381, September 1989.

\bibitem{rnn4}
{Pearlmutter}.
\newblock Learning state space trajectories in recurrent neural networks.
\newblock In {\em International 1989 Joint Conference on Neural Networks},
  pages 365--372 vol.2, 1989.

\bibitem{10.1162/neco.1997.9.8.1735}
Sepp Hochreiter and J\"{u}rgen Schmidhuber.
\newblock Long short-term memory.
\newblock {\em Neural Comput.}, 9(8):1735–1780, November 1997.

\bibitem{lstm1}
F.~A. {Gers}, J.~{Schmidhuber}, and F.~{Cummins}.
\newblock Learning to forget: continual prediction with lstm.
\newblock In {\em 1999 Ninth International Conference on Artificial Neural
  Networks ICANN 99. (Conf. Publ. No. 470)}, volume~2, pages 850--855 vol.2,
  1999.

\bibitem{lstm2}
Felix~A. Gers, Nicol~N. Schraudolph, and J\"{u}rgen Schmidhuber.
\newblock Learning precise timing with lstm recurrent networks.
\newblock {\em J. Mach. Learn. Res.}, 3(null):115–143, March 2003.

\bibitem{lstm3}
F.~A. {Gers} and E.~{Schmidhuber}.
\newblock Lstm recurrent networks learn simple context-free and
  context-sensitive languages.
\newblock {\em IEEE Transactions on Neural Networks}, 12(6):1333--1340, 2001.

\bibitem{BiLstm}
M.~{Schuster} and K.~K. {Paliwal}.
\newblock Bidirectional recurrent neural networks.
\newblock {\em IEEE Transactions on Signal Processing}, 45(11):2673--2681,
  1997.

\bibitem{dropout}
Nitish Srivastava, Geoffrey Hinton, Alex Krizhevsky, Ilya Sutskever, and Ruslan
  Salakhutdinov.
\newblock Dropout: A simple way to prevent neural networks from overfitting.
\newblock {\em J. Mach. Learn. Res.}, 15(1):1929–1958, January 2014.

\bibitem{scikit-learn}
F.~Pedregosa, G.~Varoquaux, A.~Gramfort, V.~Michel, B.~Thirion, O.~Grisel,
  M.~Blondel, P.~Prettenhofer, R.~Weiss, V.~Dubourg, J.~Vanderplas, A.~Passos,
  D.~Cournapeau, M.~Brucher, M.~Perrot, and E.~Duchesnay.
\newblock Scikit-learn: Machine learning in {P}ython.
\newblock {\em Journal of Machine Learning Research}, 12:2825--2830, 2011.

\bibitem{chollet2015keras}
François Chollet.
\newblock Keras.
\newblock \url{https://github.com/fchollet/keras}, 2015.

\bibitem{tensorflow2015-whitepaper}
Mart\'{\i}n Abadi, Ashish Agarwal, Paul Barham, Eugene Brevdo, Zhifeng Chen,
  Craig Citro, Greg~S. Corrado, Andy Davis, Jeffrey Dean, Matthieu Devin,
  Sanjay Ghemawat, Ian Goodfellow, Andrew Harp, Geoffrey Irving, Michael Isard,
  Yangqing Jia, Rafal Jozefowicz, Lukasz Kaiser, Manjunath Kudlur, Josh
  Levenberg, Dandelion Man\'{e}, Rajat Monga, Sherry Moore, Derek Murray, Chris
  Olah, Mike Schuster, Jonathon Shlens, Benoit Steiner, Ilya Sutskever, Kunal
  Talwar, Paul Tucker, Vincent Vanhoucke, Vijay Vasudevan, Fernanda Vi\'{e}gas,
  Oriol Vinyals, Pete Warden, Martin Wattenberg, Martin Wicke, Yuan Yu, and
  Xiaoqiang Zheng.
\newblock {TensorFlow}: Large-scale machine learning on heterogeneous systems,
  2015.
\newblock Software available from tensorflow.org.

\bibitem{adam}
Diederik~P. Kingma and Jimmy Ba.
\newblock Adam: {A} method for stochastic optimization.
\newblock In Yoshua Bengio and Yann LeCun, editors, {\em 3rd International
  Conference on Learning Representations, {ICLR} 2015, San Diego, CA, USA, May
  7-9, 2015, Conference Track Proceedings}, 2015.

\end{thebibliography}






\end{document}